\documentclass[10pt,twocolumn,letterpaper]{article}

\usepackage[pagenumbers]{cvpr} 

\newcommand{\smallsec}[1]{\vspace{0.2em}\noindent\textbf{#1}}

\usepackage{graphicx}

\usepackage{wrapfig}

\usepackage[dvipsnames]{xcolor}

\usepackage{url}
\usepackage{xspace}
\usepackage{caption}
\usepackage{multirow}
\usepackage{multicol}

\usepackage{tcolorbox}

\newcommand{\methodname}[0]{UniGen-AR\xspace }

\definecolor{cvprblue}{rgb}{0.21,0.49,0.74}
\usepackage[pagebackref,breaklinks,colorlinks,allcolors=cvprblue]{hyperref}

\def\paperID{15000} 
\def\confName{CVPR}
\def\confYear{2026}

\title{UniGen-AR: Unifying Visual Generation with Auto-Regressive Modeling}

\author{
Zhipeng Bao$^{1}$ \quad 
Zhen Zhu$^{2}$ \quad 
Nupur Kumari$^{1}$ \quad \\
Anurag Bagchi$^{1}$ \quad 
Yu-Xiong Wang$^{2}$ \quad 
Pavel Tokmakov$^{3}$\textsuperscript{\dag} \quad 
Martial Hebert$^{1}$\textsuperscript{\dag} \\
{\normalsize $^{1}$Carnegie Mellon University \quad $^{2}$University of Illinois Urbana-Champaign \quad $^{3}$Toyota Research Institute} \\
}

\definecolor{zhen}{rgb}{0.08, 0.38, 0.74}
\definecolor{Gray}{gray}{0.92}

\begin{document}

\twocolumn[{%
    \maketitle
    \begin{center}
        \centering
        \vspace{-20pt}
        \includegraphics[width= \linewidth]{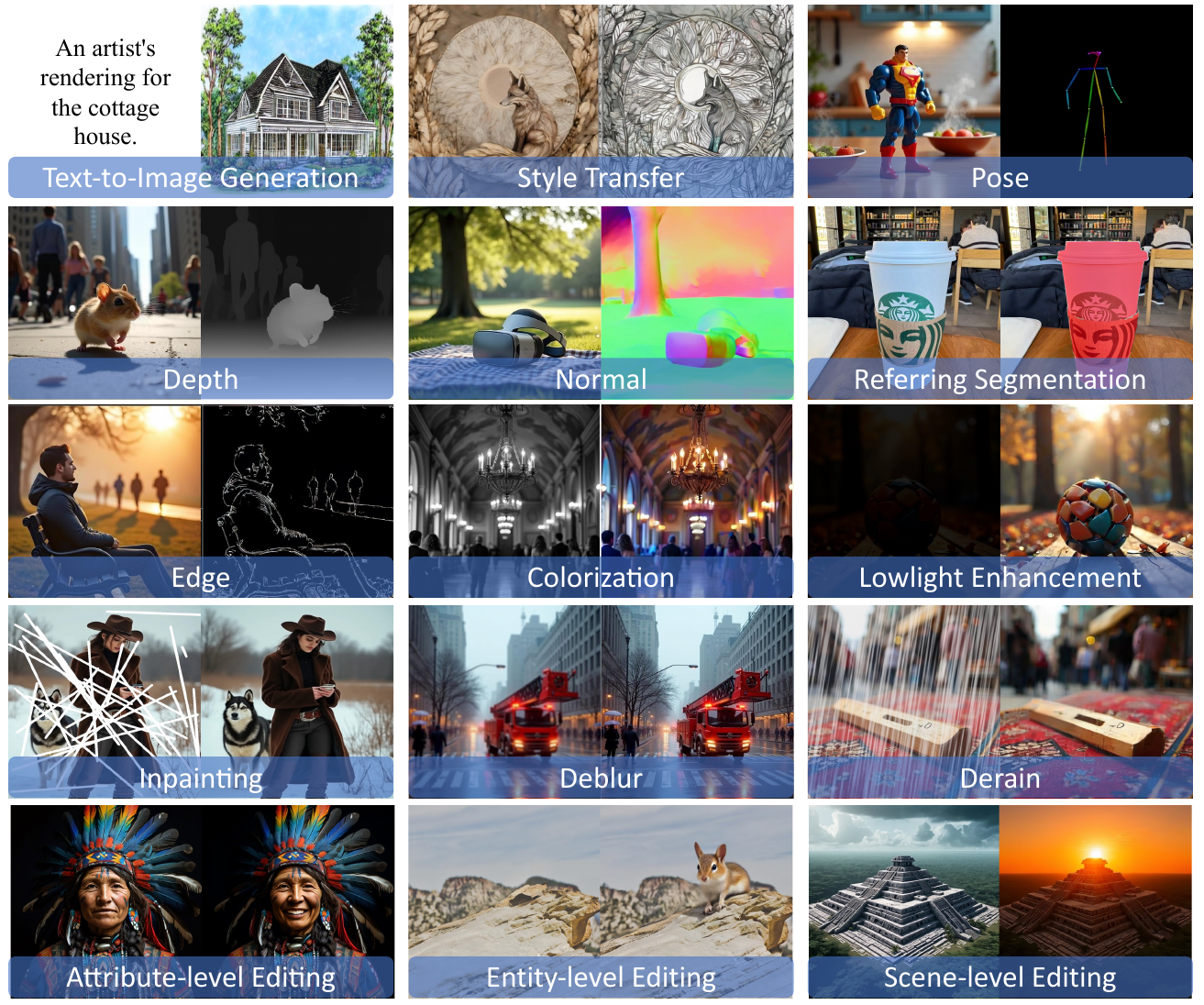}
        \vspace{-20pt}
        \captionof{figure}{\textbf{\methodname: A single model for Unified Visual Generation (UVG).} Our framework jointly handles over 15 diverse tasks, spanning text-to-image synthesis, restoration, perception, and image editing. All outputs are generated by a single, MLLM-conditioned auto-regressive backbone, using a unified prompting interface and a single set of model weights without any task-specific heads.}
        \label{fig:teaser}
    \end{center}
}]

\begin{abstract}

Modern computer vision pipelines remain fragmented, with tasks such as text-to-image generation, editing, restoration, and classical perception handled by separate models. We study Unified Visual Generation (UVG), where a single model produces diverse image-valued outputs through a unified multimodal interface. While diffusion-based systems dominate UVG due to strong quality and controllability, their iterative sampling incurs substantial inference latency, limiting practical deployment.
To address these limitations, we propose \methodname, a framework that pairs a general-purpose multi-modal language model (MLLM) with an efficient next-scale visual auto-regressive (VAR) decoder. This design retains the flexibility of MLLM-based conditioning while leveraging the sampling efficiency and latent unification properties of VAR models. In our framework, the MLLM encodes free-form instructions and control signals into a unified sequence, which guides the VAR decoder to generate image-valued outputs for over 15 tasks spanning four families.
Empirically, \methodname achieves up to $19 \times$ lower inference latency than diffusion-based baselines while maintaining or improving output quality. Our ablations further reveal that VQ-VAE tokenizer design, particularly codebook size and hierarchy, is a critical factor for VAR scalability in UVG. These results establish visual auto-regressive modeling as a compelling and efficient backbone for unified visual generation.
Our project page is at \url{https://zpbao.github.io/projects/unigenar}.

\end{abstract}
    
\section{Introduction}
\label{sec:intro}

Modern visual generation pipelines remain fragmented: text-to-image synthesis, local and global editing, restoration, and classical vision tasks, such as depth and surface normal estimation, are typically handled by separate models or specialized heads~\cite{rombach2022high,liu2025step1x,zheng2024diff,zhu2025training,bagchi2024refereverything}. Recent advances in generative modeling, however, suggest a different paradigm: many of these problems can be formulated under a shared \emph{image-out} interface, where diverse outputs are generated conditionally from visual input and optional instructions. This motivates the emerging setting of \emph{Unified Visual Generation} (UVG)~\cite{wu2025qwen,li2025visualcloze,hurst2024gpt}, in which a single model produces heterogeneous \emph{image-valued} outputs through a unified conditioning interface. Such a formulation promises shared representations, consistent controllability, and simplified deployment. 

Despite this appealing formulation, building effective UVG systems remains challenging. As the number of supported tasks grows, the model must simultaneously handle heterogeneous visual domains, diverse output formats, and multimodal control signals. This combinatorial complexity introduces substantial capacity and optimization challenges, making \emph{scalability} the central obstacle for unified visual generation. In practice, a UVG system must satisfy two key requirements: (1) a flexible multimodal interface capable of interpreting rich instructions and control signals, and (2) a scalable visual generation backbone that can efficiently produce diverse image-valued outputs.

Recent UVG systems address the first requirement by leveraging powerful multimodal large language models (MLLMs) to interpret instructions and convert heterogeneous inputs into a unified conditioning representation~\cite{bai2025qwen2}. For the generative backbone, most existing approaches adopt \emph{diffusion-based} architectures~\cite{ho2020denoising}. This combination has proven highly effective: systems such as Qwen-Image and Flux demonstrate strong generation quality and broad task coverage~\cite{batifol2025flux,wu2025qwen,comanici2025gemini}. However, diffusion models rely on iterative denoising procedures that typically require dozens of sequential sampling steps during inference. When combined with MLLM conditioning, this leads to substantial inference latency, limiting the scalability of diffusion-based UVG systems. This observation motivates the search for alternative generative backbones that retain flexible conditioning while offering a more favorable \textbf{latency--quality trade-off}.

In this work, we revisit \emph{auto-regressive} (AR) modeling as a promising backbone for UVG. While early AR models operating on coarse image tokens struggled to produce fine-grained visual details, recent advances in \emph{visual auto-regressive} (VAR) modeling show that \emph{next-scale prediction} over discrete visual tokens can achieve high-fidelity image synthesis with stable likelihood-based training~\cite{tian2024visual}. Compared with diffusion models, VAR decoders offer two attractive properties. First, they enable \emph{latent unification}: both natural images and structured outputs such as depth maps can be represented within a shared discrete token space. Second, they provide \emph{sampling efficiency}: autoregressive decoding requires substantially fewer generation stages than diffusion-based sampling while maintaining competitive perceptual quality~\cite{han2025infinity,tian2024visual}. These properties make VAR a compelling candidate for building scalable UVG systems.

However, efficient visual decoding alone is insufficient for UVG. To support rich instruction-based control, the generative backbone must be paired with a multimodal interface capable of interpreting free-form instructions and heterogeneous inputs~\cite{wu2025qwen,li2025visualcloze}. We therefore propose \textbf{\methodname}, a unified framework that combines a general-purpose MLLM encoder with a visual auto-regressive decoder. In our architecture, the MLLM parses instructions and visual inputs (\eg, text prompts and reference images) into a unified conditioning sequence, which guides a VAR-based decoder to predict discrete visual tokens. These tokens are subsequently decoded into the final image or dense-map output via a VQ-VAE~\cite{gu2022vector}. This design retains the flexible conditioning interface of recent UVG systems while benefiting from the efficiency of autoregressive decoding.

To evaluate our framework, we instantiate \methodname by repurposing a powerful pre-trained \emph{text-to-image} VAR model~\cite{han2025infinity} for the full UVG setting. We train a single backbone jointly across over \emph{15 tasks} spanning four families: text-to-image generation, classical perception, restoration, and editing. Typical visualizations are shown in Fig.~\ref{fig:teaser}. Training is performed with a unified likelihood objective under a consistent MLLM-conditioned interface. Empirically, \methodname achieves strong performance across all tasks, outperforming prior AR-based systems and exhibiting particularly strong results on restoration and classical perception benchmarks. Compared with diffusion-based UVG models under matched conditioning, \methodname achieves a favorable latency--quality Pareto frontier, reducing inference latency by up to \textbf{$\sim 5\times$} while maintaining or improving output quality. Further analysis reveals that the design of the VQ-VAE tokenizer, particularly codebook size and hierarchy, plays a critical role in scaling VAR models for UVG.

\vspace{0.5em}
\noindent\textbf{Our contributions are summarized as follows:}
\begin{itemize}
\item We introduce \textbf{\methodname}, the first framework that scales VAR modeling to the full \emph{Unified Visual Generation} setting, unifying open-ended synthesis, restoration, editing, and perception within a single image-out backbone.

\item We demonstrate strong empirical performance across more than \emph{15 tasks}, achieving on-par or superior results compared with prior UVG systems, with particularly strong improvements on restoration benchmarks.

\item We show that autoregressive decoding provides a favorable \textbf{latency--quality trade-off} compared to diffusion-based models, achieving up to \textbf{$19\times$} lower inference latency while maintaining competitive or improved quality.

\item We identify the design of the VQ-VAE tokenizer, including codebook size and hierarchical structure, as a critical factor for scaling visual autoregressive models, providing practical insights for building effective VAR-based generation systems.

\end{itemize}
\section{Related Work}
\label{sec:related}

\smallsec{Unified visual generation} aims to support tasks such as text-to-image generation, editing, restoration, and perception within a single model. Early methods explored shared latent spaces using variational autoencoders (VAEs) and vision transformers~\cite{lu2022unified,lu2024unified,kingma2013auto}. More recent approaches have adopted diffusion-based models, pretrained on large-scale data, to address a wide range of generative tasks under a unified interface~\cite{xiao2025omnigen,fu2025univg,wu2025omnigen2,han2024ace,lin2025realgeneral,brooks2023instructpix2pix,zhao2023uni,li2024brushedit}. 

A second strand builds MLLM‑mediated pipelines for controllable generation~\cite{wu2025qwen,liu2025step1x,batifol2025flux,pan2025transfer,li2025visualcloze}. Among them,  Qwen‑Image~\cite{wu2025qwen}, Step1X‑Edit~\cite{liu2025step1x}, and Metaquries~\cite{pan2025transfer} pair powerful multimodal front‑ends with diffusion decoders for text rendering and precise editing. Flux-Kontext~\cite{batifol2025flux} and Visualcloze~\cite{li2025visualcloze} specifically focus on in-context learning to enable models with the capability to learn from few-shot examples. 

A smaller yet growing body of work investigates AR modeling for image generation~\cite{sun2024x,lai2025unleashing,sun2023emu,sun2024generative,tang2024codi,ge2024seed,bai2024sequential,zhang2026nextflow}. These models typically adopt the standard ``next-token prediction'' formulation. In contrast, our work builds on the ``next-scale prediction'' paradigm introduced in visual autoregressive (VAR) models~\cite{tian2024visual}, which is better suited for UVG due to its coarse-to-fine decoding strategy.

\smallsec{Visual auto-regressive models}
factorize image generation into scale-wise predictions over discrete visual tokens, allowing coarse-to-fine decoding. The foundational work of~\cite{tian2024visual} demonstrates favorable scaling laws and superior latency–quality trade-offs compared to diffusion models. Subsequent studies extend this formulation to conditional image generation in different domains beyond ImageNet~\cite{shao2025continuous,zhuang2025vargpt,qu2025visual,chen2025visual,wang2025scalable,li2024imagefolder,luo2024open}.

Among them, two recent approaches adapt next-scale VAR to text-to-image generation~\cite{voronov2024switti,han2025infinity}. Both adopt cross-attention modules to inject text signals into the visual decoder, following a design similar to Stable Diffusion~\cite{rombach2022high}. Switti~\cite{voronov2024switti} introduces a refined attention masking strategy that restricts each token to attend only to spatially local neighbors within the current scale, improving inference speed. Infinity~\cite{han2025infinity} identifies large codebook sizes in VQ-VAE as key to achieving high-quality synthesis.
A few recent works have begun extending VAR beyond text-to-image to support editing~\cite{wang2025editinfinity,wang2025training,mao2025visual}. However, EditInfinity~\cite{wang2025editinfinity} and related methods~\cite{wang2025training} do not support direct instruction-guided UVG. Instead, they rely on indirect mechanisms such as modifying attention maps or text embeddings to steer edits. One most recent approach, VAREdit~\cite{mao2025visual}, adapts VAR models for editing, but focuses solely on this task and does not incorporate MLLMs or address broader UVG settings. In contrast, we present the first framework that combines next-scale VAR modeling with MLLM-based conditioning to support full-spectrum UVG tasks.

\smallsec{Unified models} aim to handle both understanding (text-out) and generation (image-out) in a single architecture. One direction pursues tightly-coupled token-based models~\cite{team2024chameleon,wang2024emu3,wu2024vila,jin2023unified,jin2024video,tang2025ugen,sun2024x}. 
A second direction explores hybrid designs that decouple encoding and decoding while maintaining a single interface. These approaches pair MLLM front-ends with task-specific decoders~\cite{wu2025janus,koh2023generating,pan2023kosmos,zhang2025nexus,huang2025wegen,deng2025emerging}, often sharing a central autoregressive core to support both instruction following (text-out) and controllable generation or editing (image-out). This design balances task flexibility with operational simplicity.
Our framework follows this hybrid philosophy. By coupling an MLLM front-end with a next-scale VAR decoder, we enable instruction-conditioned image generation across a wide task spectrum. Preliminary results (see the supplementary) also indicate that jointly fine-tuning the MLLM and visual decoder improves alignment between vision and language representations, suggesting a promising path toward fully unified multimodal models, which we leave for future exploration.

\section{Method}
\label{sec:method}

\begin{figure*}[t]
    \centering
    \includegraphics[width=\linewidth]{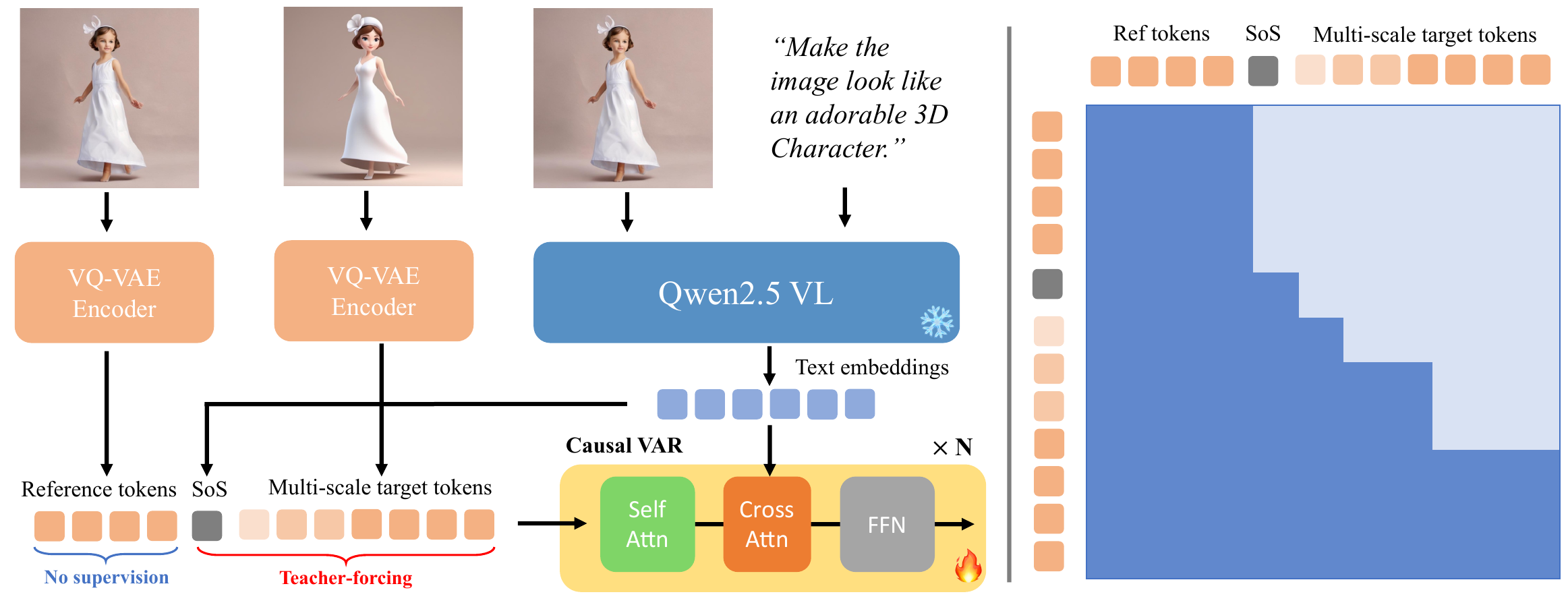}
    \vspace{-20 pt}
\caption{\textbf{Architecture of \methodname.} \textbf{Left}: 
We extend an Infinity-style VAR backbone with a multimodal language model encoder to support Unified Visual Generation with reference images. The MLLM encodes the instruction and reference image, whose text embeddings seed a learnable [SoS] token and provide keys/values for cross-attention, while finest-scale reference tokens are prepended as a non-predictive context prefix; \textbf{Right}: A block-wise causal mask lets all target tokens attend to reference and text context while preserving the standard coarse-to-fine VAR schedule.}
    \label{fig:model}
\vspace{-15pt}
\end{figure*}

This section presents the design of \methodname (illustrated in Fig.~\ref{fig:model}). We begin by reviewing VAR modeling as the core generative backbone. We then detail how we perform UVG based on existing T2I VAR models.

\subsection{Preliminary: Visual Auto-Regressive Modeling}

\smallsec{Vanilla VAR models.} VAR~\cite{tian2024visual} generates high-fidelity images by autoregressively predicting discrete visual tokens across multiple \emph{spatial scales}. It operates in a latent space defined by a multi-scale vector-quantized tokenizer, typically implemented via a VQ-VAE~\cite{kingma2013auto,van2017neural}.

Given an image $I$, the encoder $\mathcal{E}$ produces $K$ token maps:$R = \mathcal{E}(I) = (r_1, r_2, \ldots, r_K), \quad r_k \in [V]^{h_k \times w_k}$, 
where $V$ is the codebook size, and spatial resolution increases with scale index $k$ (\ie, $h_1 w_1 \leq \cdots \leq h_K w_K$). The decoder $\mathcal{D}$ reconstructs the image via:
$\hat{I} = \mathcal{D}(r_1, \ldots, r_K)$.

With the training multi-scale token sequence $R$, VAR defines an autoregressive factorization across scales:
\begin{equation}
    p_\theta(R) = \prod_{k=1}^{K}p_\theta (r_k|r_{<k}),
\label{eq:var-factorization}
\end{equation}
where $r_{<k}$ denotes tokens from coarser scales. Each $r_k$ is predicted in parallel, conditioned on $r_{<k}$ and scale-specific position embeddings. A block-wise causal mask ensures each token in $r_k$ only attends to tokens in $r_{\leq k}$. Inference proceeds sequentially from $k=1$ to $K$, with key-value caching for efficiency.

The model is trained with teacher forcing using the sum of cross-entropy losses over all token positions and scales:
\begin{equation}
    \mathcal{L}_{\text{VAR}}
= \sum_{k=1}^{K} \sum_{i=1}^{n_k}
    \text{CE}\!\left( p_\theta(\cdot \mid r_{<k}) , \; r_k^{(i)} \right),
\label{eq:var-xent}
\end{equation}
where $n_k = h_k w_k$ is the number of tokens at scale $k$, and $r_k^{(i)}$ is the ground-truth token (from $\mathcal{E}$) at position $i$.

\smallsec{Text‑conditioned next‑scale prediction.} 
To enable text-to-image generation, the VAR factorization in Equation~\eqref{eq:var-factorization} is extended to condition on a prompt $t$, encoded by a frozen language model $\psi(t)$:
\begin{equation}
p_\theta(R)=\prod_{k=1}^Kp_\theta(r_k|_{<k}, \psi(t)).
\end{equation}
Following prior work~\cite{han2025infinity,voronov2024switti}, the text embedding $\psi(t)$ is injected at each transformer layer via cross-attention. A learnable start-of-sequence ([SoS]) token, derived from a projection of $\psi(t)$, is prepended at the coarsest scale to bootstrap generation. Training remains unchanged, using block-wise causal masking and the loss in Equation~\eqref{eq:var-xent}.

Infinity~\cite{han2025infinity}, a state-of-the-art T2I VAR model serving as our backbone VAR model in our main experiments, replaces the standard categorical tokenizer with a \emph{bitwise} tokenizer that encodes each visual token as a binary vector. Instead of predicting a single index, Infinity predicts the bits of this code, so that increasing the bit-width enlarges the effective vocabulary exponentially while only mildly growing the classifier head. This binary indexing enables extremely large visual vocabularies and thus higher-fidelity reconstruction and richer visual details. The bitwise design remains compatible with cross-entropy-style training, implemented as independent binary cross-entropy losses over bits. At inference time, Infinity conditions on the text prompt $t$ and autoregressively generates the multi-scale residual token maps $(r_1, \ldots, r_K)$ with cached key--value attention states, following the VAR coarse-to-fine schedule.

\subsection{From Text-to-Image to Unified Visual Generation}

We use Infinity~\cite{han2025infinity} as our T2I backbone and extend its architecture to UVG with reference images. To enable image-to-image transformations, the model must first understand the input images; we therefore introduce a multimodal encoder that processes reference images (and text) into a shared token space. Our overall design is illustrated in Fig.~\ref{fig:model}.

\smallsec{Reference- and target-tokenization.}
Given a reference image $I^{\mathrm{ref}}$ and a target image $I^{\mathrm{tar}}$, we first encode them with the same multi-scale VQ-VAE encoder $\mathcal{E}$:
\begin{equation}
    R^{\mathrm{ref}} = \mathcal{E}(I^{\mathrm{ref}}), \quad 
    R^{\mathrm{tar}} = \mathcal{E}(I^{\mathrm{tar}}).
\end{equation}
For the target image, we keep all scales
$R^{\mathrm{tar}} = (r^{\mathrm{tar}}_1,\dots,r^{\mathrm{tar}}_K)$ as in vanilla Infinity. For the reference image, we only retain the finest scale
$r^{\mathrm{ref}}_K$, and discard coarser scales:
\begin{equation}
    R^{\mathrm{ref}} = (r^{\mathrm{ref}}_K), \quad r^{\mathrm{ref}}_K \in [V]^{h_K \times w_K}.
\end{equation}
The concurrent work, EditVAR~\cite{mao2025visual}, also demonstrates that finest-scale tokens are sufficient for UVG tasks. This design, prepending the finest reference tokens, leaves the original VAR scale schedule unchanged: the reference tokens are never traversed by the next-scale generation process and are excluded from the loss.

\smallsec{Multimodal encoder.}
We replace the Infinity's text encoder, T5~\cite{chung2024scaling}, with a multimodal language model $\psi$ (Qwen2.5-VL~\cite{bai2025qwen2}) to encode the input instruction $t$, together with the reference image: $Z_t = \phi(I^\text{ref}, t)$. Similar to prior efforts in the Diffusion regime~\cite{wu2025qwen}, $Z_t$ are the \emph{pure text embeddings} from the last self-attention layer of the MLLM encoder to remove the redundancy. These text embeddings are used in two ways, following Infinity~\cite{han2025infinity}: a learnable [SoS] token is obtained by projecting the text embedding and prepended at the coarsest scale to bootstrap generation, and the text embedding also serves as the key and value sequence for the cross-attention layers that modulate the visual tokens.

\smallsec{Unified token sequence and causal masking.}
We unify reference and target tokens into a single autoregressive sequence
\begin{equation}
    \mathbf{s} = \big( {r}^{\mathrm{ref}},\ \text{[SoS]},\ {r}^{\mathrm{tar}}_1,\dots,{r}^{\mathrm{tar}}_K \big),
\end{equation} 
where the reference tokens are \emph{prepended} before the [SoS] token. Conceptually, $r^{\mathrm{ref}}$ acts as a non-predictive context prefix, the [SoS] token marks the autoregressive start, and the multi-scale target tokens follow the standard VAR schedule.

We implement a block-wise causal mask $M \in \{0,1\}^{|\mathbf{s}|\times|\mathbf{s}|}$, visualized on the right side of Fig.~\ref{fig:model}, that enforces:
\begin{itemize}
    \item reference tokens are visible to all subsequent tokens ([SoS] and target tokens) but are never used as prediction targets;
    \item the [SoS] token and all target tokens respect the original next-scale causal ordering: tokens in scale $k$ can attend to reference tokens, [SoS], and all tokens in $r^{\mathrm{tar}}_{<k}$, but not to future scales.
\end{itemize}

\smallsec{Training and inference.}
Training follows teacher forcing as in Equation~\eqref{eq:var-xent}. At inference time, we perform iterative next-scale prediction as in standard VAR: we first sample $r^{\mathrm{tar}}_1$ conditioned on $\mathbf{r}^{\mathrm{ref}}$ and $t$, then proceed to finer scales until $r^{\mathrm{tar}}_K$ is obtained. Due to the causal mask, tokens at each step can freely attend to the entire reference sequence and the text context while respecting the multi-scale ordering. When no reference image is provided, the same decoding procedure reduces to conventional T2I generation.

\section{Experimental Evaluations}
\label{sec:exp}

\begin{table*}[t]
    \centering
    \resizebox{\linewidth}{!}{
    \begin{tabular}{l|c|cccc|ccc}
    \toprule 
       \multirow{2}{*}{Model} & \multirow{2}{*}{\# Params} & \multicolumn{4}{c|}{GenEval} & \multicolumn{3}{c}{Emu-Edit} \\ 
       \cmidrule(lr){3-9}
       & & Two Obj. ($\uparrow$) & Position ($\uparrow$) & Color Attr. ($\uparrow$) & Overall ($\uparrow$) & CLIP-I ($\uparrow$) & CLIP-OUT ($\uparrow$) & DINO ($\uparrow$) \\ \midrule
       SDv2.1~\cite{rombach2022high} & 0.9B & 0.51 & 0.07 & 0.17 & 0.50 & - & - & -\\
       DALL-E 2~\cite{ramesh2022hierarchical} & 6.5B & 0.66 & 0.10 & 0.19 & 0.52 & - & - & -\\
       DALL-E 3~\cite{betker2023improving} & - & - & - & - & 0.67 & - & - & -\\
       SDXL~\cite{podell2023sdxl} & 2.6B & 0.74 & 0.15 & 0.23 & 0.55 & - & - & -\\ 
       SD3 (d=21)~\cite{esser2024scaling} & 2B & 0.74 & 0.34 & 0.36 & 0.62 & - & - & -\\
       InstructPix2Pix~\cite{brooks2023instructpix2pix} &0.9B & - & - & - & - & 0.824 & 0.268 & 0.765 \\
       Step-1X~\cite{liu2025step1x} & 12B & - & - & - & - &  0.831 &  \underline{0.275} & 0.773 \\
       OmniGen2~\cite{wu2025omnigen2} & 4B & \bf 0.95 & \bf 0.55 & \bf 0.76 & \bf 0.80 & \bf 0.876 & \bf 0.309 & \bf 0.822 \\ 
       \midrule
       LlamaGen~\cite{sun2024autoregressive} & 0.8B & 0.34 & 0.07 & 0.04 & 0.32  & - & - & -\\
       Chameleon~\cite{team2024chameleon} & 7B & - & - & - & 0.39  & - & - & -\\
       Emu3~\cite{wang2024emu3} & 8.5B & 0.81 & \underline{ 0.49} & 0.45 & 0.66  & - & - & - \\
       Infinity~\cite{han2025infinity} & 2B & \underline{ 0.85} & \underline{0.49} & \underline{ 0.57} & \underline{ 0.73}  & - & - & - \\
       VAREdit~\cite{mao2025visual} & 8B & - & - & - & - & - & 0.271 & - \\
       
          \midrule 
          Infinity (UVG) & 2B & 0.80 & 0.38 & 0.48 & 0.70 & 0.827 & 0.270 & 0.759 \\ 
         \methodname (Ours) & 2B & 0.76 & 0.41 & 0.45 & 0.68 & \underline{ 0.849} & 0.273 & \underline{ 0.791}\\ 
         
         \bottomrule
    \end{tabular}
    }
    \caption{\textbf{Comparison on T2I generation and image editing.} 
    Top: diffusion-based models; Bottom: autoregressive models. 
\methodname achieves competitive performance while supporting over 15 visual generation tasks and being trained using only public data. \# Params excludes text encoders.
Infinity (UVG) denotes the version of Infinity finetuned for UVG using the same configuration as \methodname.
}
    \label{tab:main-t2i}
    \vspace{-5pt}
\end{table*}

\begin{table*}[t]
    \centering
    \resizebox{\linewidth}{!}{
    \begin{tabular}{l|cc|cccccc}
    \toprule 
    \multirow{2}{*}{Model} & NYUv2-Depth & NYUv2-Normal & \multicolumn{2}{c}{LOL-Lowlight} & \multicolumn{2}{c}{GoPro-Deblur} & \multicolumn{2}{c}{Rain100L-Derain} \\ 
    & RMSE ($\downarrow$) & Mean Angle Err ($\downarrow$) & PSNR ($\uparrow$) & SSIM ($\uparrow$) & PSNR ($\uparrow$) & SSIM ($\uparrow$) & PSNR ($\uparrow$) & SSIM ($\uparrow$) \\ \midrule
       Depth Anything~\cite{yang2024depth} & \color{gray} 0.206 & - & - & - & - & - & - & -\\
       Marigold~\cite{ke2024repurposing} & \color{gray} 0.224 & - & - & - & - & - & - & -\\ 
       Bae \etal~\cite{bae2021estimating} & - & \color{gray} 14.90 & - & - & - & - & - & - \\
       InvPT~\cite{ye2022inverted} & - & \color{gray} 19.04 & - & - & - & - & - & - \\
       AirNet~\cite{li2022all} & - & - & \color{gray} 18.18 & \color{gray} 0.735 & \color{gray} 24.35 & \color{gray} 0.781 & \color{gray} 32.98 & \color{gray} 0.951\\
       InstructIR~\cite{conde2024instructir} & - & - & \color{gray} 23.00 & \color{gray} 0.836 & \color{gray} 29.40 & \color{gray} 0.886 & \color{gray} 36.84 & \color{gray} 0.937\\ \midrule
       
       InstructCV~\cite{gan2023instructcv} & 0.297 & - & - & - & - & - & - & - \\
       UnifiedIO~\cite{lu2022unified} & 0.387 & - & - & - & - & - & - & -\\
        OmniGen~\cite{xiao2025omnigen} &  0.480 & - & 13.38 & 0.392 & 13.39 & 0.321 & 12.02 & 0.233\\
        Painter~\cite{wang2023images} & 0.288  & - & \bf 22.40 & \bf 0.872 &  - & - & 29.87 & 0.882\\ 
       VisualCloze~\cite{li2025visualcloze} & 0.285 & 20.98 & 18.53 & 0.796 & 19.75 & 0.691 & 23.27 & 0.780 \\ 
        
        X-Prompt~\cite{sun2024x} & 0.277 & 19.17 &  19.71 & 0.810 & 21.04 & 0.761 & 25.53 & 0.843\\ \midrule 
        Infinity (UVG) & 0.261 & 19.09 & 20.10 & 0.813 & 22.17 & 0.758 & 29.32 & 0.886 \\
        \methodname (Ours) & \bf 0.245 & \bf 18.76 & 21.03 & 0.825 & \bf 22.99 & \bf 0.774 & \bf 33.71 & \bf 0.926 \\ 
         \bottomrule
    \end{tabular}
    }
    \caption{\textbf{Results on perception and restoration tasks.} Top: specialized models; Bottom: recent UVG systems. \methodname significantly outperforms AR-based UVG prior work (X-Prompt) and demonstrates competitive performance on image restoration tasks, even compared to the specialized models. }
    \label{tab:main-cv}
    \vspace{-10pt}
    
\end{table*}

\subsection{Experimental Setup}

\smallsec{Training data.} 
Following prior work~\cite{li2025visualcloze,xiao2025omnigen,wei2024omniedit}, we train \methodname using publicly available paired datasets. For text-to-image (T2I) generation, we use the LAION-COCO-Aesthetic subset~\cite{xiao2025omnigen,schuhmann2021laion}, containing approximately 4M images. For perception tasks, we adopt the Graph200K dataset from VisualCloze~\cite{li2025visualcloze}, which provides 200K images paired with annotations for depth estimation, surface normals, edge detection, and human pose estimation. For image restoration, we consider five tasks: deblurring, deraining, colorization, inpainting, and low-light enhancement. We follow the processing scheme of VisualCloze to generate the noisy version of these labels. For referring image segmentation, we use RefCOCO~\cite{yu2016modeling}, which contributes roughly 320K (image, mask, ref\_prompt) triplets. For the whole set of editing tasks, we leverage OMNI-EDIT~\cite{wei2024omniedit} as our training source, with 1.2M images and 2M instructions. We additionally merge StyleBooth~\cite{han2025stylebooth} into our datamix to enhance style transfer. 
In total, our training corpus comprises roughly 8M paired examples across over 15 UVG tasks.

\smallsec{Implementation details.} 
We initialize our system from the pretrained Infinity-2B model~\cite{han2025infinity}. We replace its original T5~\cite{chung2024scaling} text encoder with Qwen2.5-VL (3B)~\cite{bai2025qwen2} for multimodal conditioning. Following~\cite{wu2025qwen}, we extract only the textual embeddings from the MLLM and omit visual embeddings.
Training proceeds in two stages:
\textbf{Stage I (alignment).} We freeze the entire Qwen2.5-VL and Infinity backbones, except for the text-normalization layer, text-projection layer, and unconditional embeddings used for classifier-free guidance. This aligns the pretrained Infinity decoder with the new Qwen2.5-VL conditioning. In this stage, we only train the model with the T2I data. 
\textbf{Stage II (UVG training).} We jointly train the Infinity backbone components on the full datamix. Each batch is either a pure T2I batch or a mixed batch drawn from all other tasks, with the T2I sampling probability set to 0.25. All outputs are generated at a fixed resolution of $512 \times 512$.

Stage I is trained for 2 epochs with an effective batch size of 256; Stage II is trained for 100K steps with an effective batch size of 128. We use AdamW~\cite{loshchilov2017decoupled} with learning rates of 1e-4 (Stage I) and 5e-6 (Stage II). Training requires $\sim$3 days for Stage I and $\sim$7 days for Stage II on a single NVIDIA H100 node.

\begin{figure}[t]
    \centering
    \includegraphics[width=\linewidth]{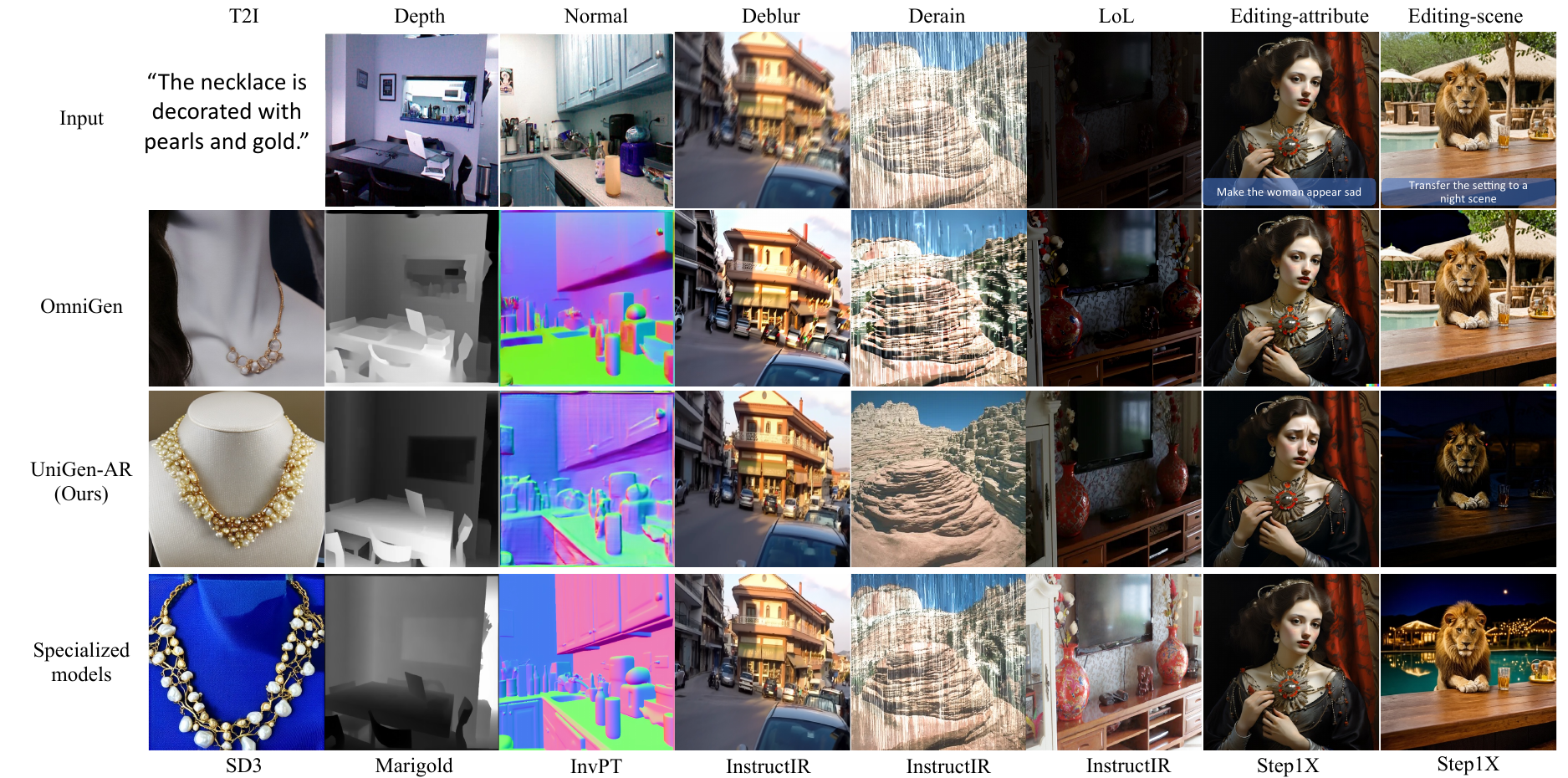}
    \vspace{-15pt}
    \caption{\textbf{Qualitative comparisons across UVG tasks.}
\methodname consistently produces higher-quality results than the UVG baseline OmniGen and achieves visually comparable or superior outputs to specialized models, notably outperforming InstructIR~\cite{conde2024instructir} on deraining.}
    \label{fig:comparison}
    \vspace{-15pt}
\end{figure}

\smallsec{Evaluation benchmarks.} We evaluate on four groups of tasks.  
\textbf{T2I generation:} GenEval benchmark~\cite{ghosh2023geneval}, following Infinity~\cite{han2025infinity}.  
\textbf{Image editing:} Emu-Edit benchmark~\cite{sheynin2024emu}, which assesses different types of editing-related tasks, including style transfer.  
\textbf{Perception tasks:} depth estimation and surface normals on NYUv2~\cite{silberman2012indoor}.  
\textbf{Restoration tasks:} low-light enhancement on LOL~\cite{wei2018deep}, deblurring on GoPro~\cite{nah2017deep}, and deraining on Rain-13K~\cite{fu2017removing}. 

Qualitative comparisons for all tasks appear in Figures~\ref{fig:teaser} and \ref{fig:comparison}, with additional examples provided in the supplementary. More details about our training data and implementation are also included in the supplementary material. 

\subsection{Main results}

We report results on T2I generation and image editing in Table~\ref{tab:main-t2i}, and results on perception and image restoration tasks in Table~\ref{tab:main-cv}. Representative outputs are shown in Fig.~\ref{fig:comparison}. For T2I and editing, we compare against both diffusion-based and autoregressive models; for perception and restoration tasks, we compare with specialized task-specific models as well as recent UVG systems. 

We additionally introduce a variant of our model as a baseline, denoted as \textit{Infinity (UVG)}. This model retains the original Infinity parameters and backbone architecture, while undergoing the same architectural modifications as \methodname\ (\eg, the introduction of reference tokens). It is then trained using the same data and training configuration as our model, enabling a controlled comparison that isolates the impact of our design.

\smallsec{Text-to-image generation and image editing.}
Table~\ref{tab:main-t2i} shows that:
(1) \methodname achieves strong performance on GenEval, outperforming larger diffusion models such as DALL-E 2~\cite{ramesh2022hierarchical}, demonstrating that next-scale VAR remains competitive even in the unified setting.  
(2) Compared with the Infinity checkpoint, our model shows a mild drop in performance, likely due to Infinity's use of large-scale proprietary training data.  
(3) Despite using only public data, our model surpasses diffusion counterparts with similar model sizes, including SD3 (2B)~\cite{esser2024scaling}. This suggests that VAR-based backbones retain strong prior knowledge during finetuning and remain a compelling alternative to diffusion for controllable image generation.

For image editing, \methodname also demonstrates strong performance. 
Compared with the recent VAR-based editing model VAREdit~\cite{mao2025visual}, our method achieves slightly better results across the editing metrics, highlighting the effectiveness of our unified training strategy.
Moreover, despite using a significantly smaller backbone (12B \vs 2B parameters), \methodname outperforms the diffusion-based strong editing baseline, Step-1X~\cite{liu2025step1x}. These results suggest that VAR-based backbones remain highly competitive for instruction-guided image editing even in the unified multi-task setting.

\smallsec{Perception and image restoration tasks.} 
Based on the results in Table~\ref{tab:main-cv}, we offer the following observations: 
(1) \methodname consistently outperforms the strongest AR-based UVG model X-Prompt~\cite{sun2024x} across all evaluated tasks, highlighting the advantage of coarse-to-fine refinement in next-scale prediction.  
(2) Compared with specialized task-specific models, a performance gap remains, which reflects the inherent challenge of UVG, where a single model must master diverse, heterogeneous objectives.  
(3) Notably, our model showcases superior performance on the restoration tasks. In particular, for low-light enhancement and derain tasks, our model surpasses a dedicated restoration model (AirNet~\cite{li2022all}), suggesting that the bitwise VQ-VAE used in Infinity provides a favorable structure for correcting token-level degradations.

\begin{table}[t]
    \vspace{-10pt}
    \centering
    \resizebox{\linewidth}{!}{
    \begin{tabular}{l|ccc|c}
    \toprule
       \textbf{Table I}  & CLIP-I & CLIP-OUT & DINO & FLOPs \\
    \midrule
       Qwen + SD3 (512)  & - & - & - & 16T \\ 
       \methodname (512) & 0.849 & 0.273 & 0.791 & 1.9T \\ 
    \midrule
       Qwen + SANA (512) & 0.831 & 0.275 & 0.724 & 14T \\
    \midrule 
       Qwen + SD3 (1024)  & - & - & - & 79T \\ 
       \methodname (1024) & \bf 0.862 & \bf 0.289 & \bf 0.810 & 4.1T \\ 
    \bottomrule
    \end{tabular}
    }
    \caption{\textbf{FLOPs per generated image comparisons with Diffusion models.} Under the same MLLM-conditioned setting, UniGen-AR requires substantially fewer FLOPs (\textbf{$>10\times$}) than the diffusion counterpart.}
    \label{tab:sana}
\end{table}

\begin{figure*}[t]
\begin{minipage}{0.57 \linewidth}
    \centering
    \resizebox{\linewidth}{!}{
    \begin{tabular}{l|c|c}
    \toprule 
       Model & Qwen + SD3 & Qwen + Infinity \\ \midrule 
       \# of Params & 3B+2B & 3B+2B \\ \midrule
       Two Obj. & 0.68 & 0.75\\
       Position &  0.36 & 0.45\\
       Color Attr. & 0.29 & 0.43 \\
       Overall & 0.52 & 0.64 \\\midrule
       Latency (s/img) & 5.23 & 1.05 \\ 
    \bottomrule
    \end{tabular}
    }
    \captionof{table}{\textbf{Diffusion \vs VAR decoders.} We compare SD3 (diffusion) and Infinity (VAR) under identical finetuning settings. VAR provides better generation accuracy and is substantially faster at inference. }
    \label{tab:var_vs_diff}
\end{minipage}
\hfill
\begin{minipage}{0.39 \linewidth}
    \centering
    \includegraphics[width=\linewidth]{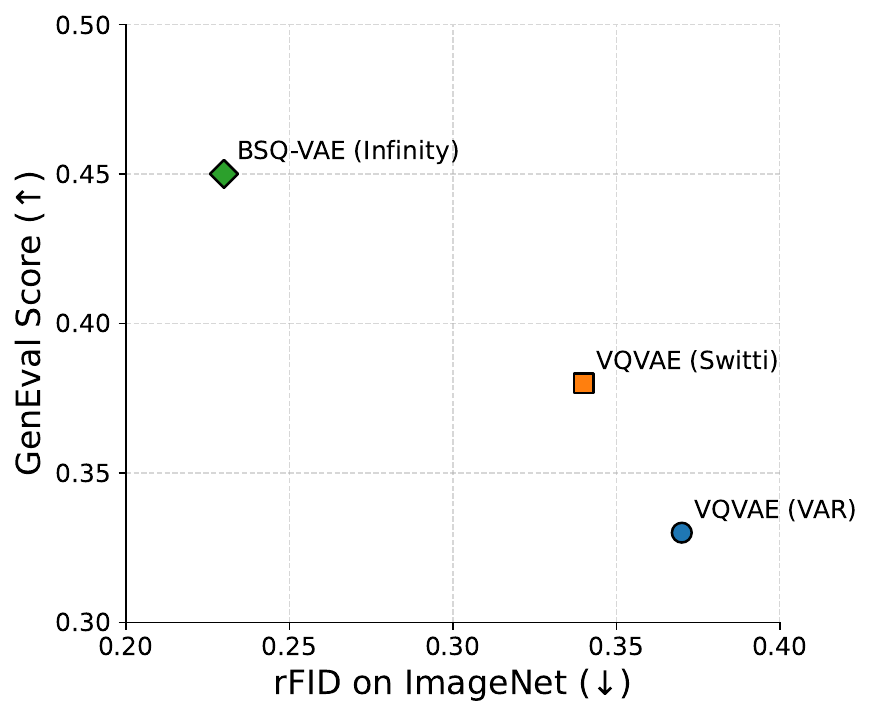}
    \vspace{-22pt}
    \caption{\textbf{Impact of visual tokenizers.} Better reconstruction fidelity leads to improved generation quality.}
    \label{fig:vqvae}
\end{minipage}
\vspace{-5pt}
\end{figure*}

\begin{table*}[t]
    \centering
    \resizebox{\linewidth}{!}{
    \begin{tabular}{l|cccc|ccc}
    \toprule
        Model & Two Obj. ($\uparrow$) & Position ($\uparrow$) & Color Attr. ($\uparrow$) & Overall ($\uparrow$) &  LOL-Lowlight ($\uparrow$) & GoPro-Deblur ($\uparrow$) & Rain100L-Derain ($\uparrow$) \\
    \midrule 
        UniGen-AR (Single) & 0.68 & 0.31 & 0.40 & 0.59 & 20.51 & 22.18 & 32.26 \\
        UniGen-AR (TwoStage) & \bf 0.76 & \bf 0.41 & \bf 0.45 & \bf 0.68 & \bf 21.03 & \bf 22.99 & \bf 33.71\\
    \bottomrule
    \end{tabular}
    }
    \caption{\textbf{Single-stage training \vs two-stage training.} The single-stage variant consistently lags behind. These results indicate that the alignment stage is necessary to adapt the pretrained VAR decoder to the MLLM's conditioning distribution.}
    \label{tab:stages}
    \vspace{-10pt}
\end{table*}

\smallsec{Qualitative comparisons.}
Fig.~\ref{fig:comparison} illustrates that \methodname produces higher-quality results than the UVG baseline OmniGen~\cite{xiao2025omnigen} across all evaluated tasks. Moreover, for most tasks, our outputs are visually comparable to those of specialized models. Notably, on the deraining task, \methodname achieves cleaner rain removal than the state-of-the-art dedicated model InstructIR~\cite{conde2024instructir}. These qualitative results further highlight the strength of our MLLM–VAR architecture for UVG and suggest promising potential for real-world applications.

\smallsec{Impact of the multimodal encoder.}
By comparing the results of Infinity (UVG) and \methodname in Tables~\ref{tab:main-t2i} and~\ref{tab:main-cv}, we observe that the variant equipped with the pure T5 text encoder achieves a slightly higher GenEval score than our default model using the Qwen encoder. However, the Qwen-based model consistently outperforms the T5 variant on the remaining tasks, particularly those that require richer visual conditioning, such as image editing and restoration. We interpret this as a \textbf{trade-off}: while T5 is slightly better aligned with the GenEval benchmark, the multimodal Qwen encoder provides stronger representations for information-dense tasks where visual context is essential. We hypothesize that this advantage arises because the MLLM encoder jointly processes the reference image and the instruction prompt, producing embeddings that implicitly encode object- and region-level semantics, which helps better handle localized reasoning required by many UVG tasks.

\subsection{Ablation Study}

\smallsec{Diffusion \vs VAR.}
We first compare the impact of the decoder architecture by replacing the Infinity VAR decoder with the SD3 diffusion decoder, while keeping all other training settings (\ie, data, steps, resolution) identical. For efficiency, all variants are finetuned for two epochs on the T2I subset only; therefore, absolute numbers differ from Table~\ref{tab:main-t2i}. Results are summarized in Table~\ref{tab:var_vs_diff}. 

Under this controlled setting, the VAR-based Infinity decoder consistently outperforms the diffusion-based SD3 decoder across all GenEval categories. These results suggest that next-scale prediction provides a competitive alternative to diffusion. Beyond generation quality, the two architectures exhibit markedly different inference characteristics. The Infinity variants achieve approximately \textbf{5$\times$ lower inference latency} than their SD3 counterparts under identical hardware and resolution settings, owing to their efficient autoregressive decoding stages. 

Since wall-clock latency can be affected by implementation-specific optimizations, we additionally report FLOPs per generated image in Table~\ref{tab:sana} for both 512 and 1024 resolutions. Under the same MLLM-conditioned setting, UniGen-AR requires substantially fewer FLOPs (\textbf{$>10\times$}) than the diffusion counterpart. Moreover, we also train a Qwen+SANA~\citep{xie2024sana} baseline, an efficient diffusion model, for image editing. With the results shown in Table~\ref{tab:sana}, UniGen-AR consistently outperforms this variant, with \textbf{~7$\times$} lower FLOPs. 

Taken together, these results indicate that VAR decoders offer a more favorable \textbf{quality--latency trade-off}, making them particularly attractive for latency-sensitive or interactive unified visual generation systems.

\smallsec{Impact of visual tokenizer.} 
We further study the influence of the discrete visual tokenizer by training three variants using VQ-VAEs from VAR~\cite{tian2024visual}, Switti~\cite{voronov2024switti}, and Infinity~\cite{han2025infinity}. Different from the core experiment, here we finetune the checkpoint from the \textbf{vanilla VAR model}
Both models are trained in a single-stage T2I setting for two epochs. Fig.~\ref{fig:vqvae} shows reconstruction FID (rFID) on ImageNet~\cite{deng2009imagenet,Heusel2017gans} and GenEval~\cite{ghosh2023geneval} performance. We observe a strong inverse correlation between rFID and generation quality: tokenizers with lower reconstruction error yield higher GenEval scores. This underscores the visual tokenizer as a central bottleneck in VAR-style architectures. Improving token expressiveness and reconstruction fidelity remains a promising future research direction.

\begin{table*}[t]
\vspace{-10pt}
    \centering
    \begin{tabular}{l|cccc}
    \toprule
        \textbf{Model Variant} & Two Obj. ($\uparrow$) & Position ($\uparrow$) & Color Attr. ($\uparrow$) & Overall ($\uparrow$) \\
    \midrule
        Infinity & \color{gray} 0.85 & \color{gray} 0.49 & \color{gray} 0.57 & \color{gray} 0.73 \\ \midrule
        Infinity (UVG), Stage I &  0.79 & \bf 0.40 & 0.45 & 0.68\\ 
        Infinity (UVG), Stage I$+$II & \bf 0.80 & 0.38 & \bf 0.48 & \bf 0.70 \\ 
    \bottomrule
    \end{tabular}
    \caption{ \textbf{Impact of data and multi-task training.} 
 Training data scale and quality account for a substantial portion of the gap from the Stage I results to the original Infinity. After stage-II training with the full multi-task objective, the GenEval score improves again, highlighting the synergy of multi-task learning under UVG.
 }
    \label{tab:analysis}
\end{table*}

\begin{table*}[t]
    \centering
    \begin{tabular}{l|cccc}
    \toprule 
       Setting & Two Obj. & Position & Color Attr. & Overall \\ \midrule
       Ours ((UVG only)) & 0.75 & 0.45& 0.43 & 0.64\\ 
       Ours (Joint MMU+UVG) & \bf 0.82 & \bf 0.47 & \bf 0.46 & \bf 0.69 \\
    \bottomrule
    \end{tabular}
    \caption{\textbf{Multimodal understanding improves unified visual generation.} Jointly finetuning Qwen2.5-VL for multimodal understanding leads to consistent performance gains on GenEval, especially for multi-object reasoning, highlighting the practicality of coupling understanding with generation.}
    \label{tab:mmgeneration}
    \vspace{-10pt}
\end{table*}

\smallsec{Impact of two-stage training.}  We additionally conduct an ablation on our two-stage training scheme by building a variant of our model that omits the Stage I alignment. In Table~\ref{tab:stages}, we observe consistently degraded performance across tasks under this setting. These results indicate that the alignment stage is necessary to adapt the pretrained VAR decoder to the MLLM's conditioning distribution, consistent with the standard practice in training MLLMs~\cite{liu2023visual}.

\smallsec{Impact of data and multi-task joint training.} 
In Tables~\ref{tab:main-t2i} and~\ref{tab:main-cv}, we observe that the finetuned baseline \textit{Infinity (UVG)} exhibits a noticeable drop in T2I performance compared with the original Infinity. To better isolate the contributing factors, we additionally report the GenEval score of the stage-I checkpoint in Table~\ref{tab:analysis}. By comparing these results, we find that the GenEval score decreases from $0.73$ to $0.68$ after stage I training, indicating that \textbf{training data scale and quality} account for a substantial portion of the gap to the reported Infinity number. Such degradation is consistent with prior findings showing that fine-tuning large pretrained models on smaller or distribution-shifted datasets can lead to performance drops due to catastrophic forgetting or overfitting~\cite{zhu2024model,sanyal2025upweighting}.

Interestingly, after stage-II training with the full multi-task objective, the GenEval score improves again compared with the stage-I checkpoint. This suggests that joint training across heterogeneous UVG tasks provides complementary supervision signals that benefit the shared visual representation. Such improvements highlight the \textbf{synergy of multi-task learning}, supporting the spirit of unified visual generation model design.

\subsection{Extension: Multimodel Understanding and Generation}

Thus far, we have focused on how an MLLM serves as a strong conditioner for the VAR backbone. Now we take a different perspective: treating \methodname as a visual-output branch of an MLLM, we investigate whether improving the MLLM's understanding ability can, in turn, enhance its visual generation.

To control for data and isolate the effect, we exclusively reuse the same T2I subset employed for training the generator. Each T2I sample is repurposed into a VQA-style instance by converting the caption into an answer and assigning a fixed question template: \emph{``Generate a caption for this image.''} To increase linguistic diversity, we follow and pre-sample 50 paraphrased variants of this question via Qwen2.5-VL itself for training. During joint training, we finetune the last 10 layers of Qwen2.5-VL together with the Infinity decoder. The multimodal-understanding loss updates only the Qwen layers, whereas the UVG loss updates both Qwen and Infinity in a coupled manner. Both variants are trained for two epochs on T2I data for fair comparison.

Table~\ref{tab:mmgeneration} reports the results. We observe that jointly training for multimodal understanding consistently improves T2I generation. The gains are especially pronounced for the \emph{Two Objects} category, which requires resolving relationships across multiple entities -- an ability naturally strengthened by the auxiliary understanding objective. These findings suggest that multimodal understanding and multimodal generation are mutually beneficial: enhancing the semantic reasoning capability of the MLLM leads to improved visual generation fidelity. This synergy points toward a promising direction for future unified architectures.

\subsection{Limitation and Future Work}
\smallsec{Limitation.} A key limitation of our current design lies in its fixed output resolution of $512\times512$. While this choice simplifies training, it prevents \methodname from flexibly adapting to inputs of arbitrary size. Adopting the dynamic‑resolution strategies used in prior work~\cite{rombach2022high,han2025infinity}, \eg, multiple groups of resolution choices and spatial padding, represents a practical next step toward broad deployment.
We additionally present typical \textbf{failure modes} in the supplementary.

We have two \textbf{future research directions} following: 
First, inspired by recent advances in MLLMs~\cite{liu2023visual,liu2024improved} and diffusion transformers~\cite{esser2024scaling,batifol2025flux}, we aim to move from cross‑attention conditioning toward a unified self‑attention architecture, which we believe offers stronger coupling between modalities and improved controllability. Meanwhile, our preliminary findings suggest that joint training of the MLLM and VAR decoder benefits multimodal alignment; we therefore see fully unified modeling -- capable of both multimodal understanding (text‑out) and generation (image‑out) -- as an exciting long‑term goal.

\section{Conclusion}
\label{sec:future}

We present \methodname, a unified visual generation framework that combines a multimodal language model encoder with a visual auto-regressive decoder. Our approach extends next-scale VAR modeling to the unified visual generation (UVG) setting, enabling a single backbone to handle diverse tasks including text-to-image generation, image editing, perception, and restoration. Experiments show that \methodname achieves competitive performance across these tasks while offering a favorable latency–quality trade-off compared to diffusion-based systems, highlighting the potential of autoregressive visual modeling as a scalable alternative for unified visual generation.

\section*{Acknowledgement}

This work was supported by Toyota Research Institute.


\bibliographystyle{ieeenat_fullname}
\bibliography{main}

\clearpage
\newpage
\appendix
\setcounter{figure}{0}
\setcounter{table}{0}
\setcounter{equation}{0}
\renewcommand{\thefigure}{\Alph{figure}}
\renewcommand{\thetable}{\Alph{table}}
\renewcommand{\theequation}{\Alph{equation}}
In this supplementary material, we first detail the training data of our \methodname in Section~\ref{suppsec:data}. 
Next, we present typical failure models of our \methodname in Section~\ref{suppsec:failure}. 
Then, in Section~\ref{suppsec:implementation} and Section \ref{suppsec:rvos},  we provide additional implementation details and evaluations, respectively. Finally, we provide additional visualizations for each task we considered in Section~\ref{suppsec:visual}.

\section{Training Data}
\label{suppsec:data}

We follow prior unified-vision works~\cite{li2025visualcloze,xiao2025omnigen} in constructing a large multi-task training corpus. Table~\ref{tab:supp_data} summarizes all datasets used for training \methodname. For image restoration tasks, we adopt the same synthetic pipeline as VisualCloze~\cite{li2025visualcloze}. Below, we provide task-specific details in Table~\ref{tab:supp_data}, and disclose necessary details for some of them.

\begin{table*}[t]
    \centering
    \resizebox{\linewidth}{!}{
    \begin{tabular}{l|ccc}
    \toprule
    Task & Dataset & Data Volume & Annotation \\ \midrule
T2I & Laion-coco-aesthetics~\cite{schuhmann2021laion} & 4.1m & CLIP captions \\ 
\midrule
Depth estimation & Graph200k~\cite{li2025visualcloze} & 205k & Depth anything \\
Pose estimation &  Graph200k~\cite{li2025visualcloze} & 205k & Open-pose \\
Normal prediction & Graph200k~\cite{li2025visualcloze} & 205k & DSINE \\
Edge detection & Graph200k~\cite{li2025visualcloze} & 205k & Canny \\
Referring segmentation & RefCOCO~\cite{yu2016modeling} & 320k & Human label \\
\midrule
Colorization & Graph200k~\cite{li2025visualcloze} & 205k & RGB2Gray \\
Denoising & Graph200k~\cite{li2025visualcloze} & 205k & 10 random noise \\
Inpainting & Graph200k~\cite{li2025visualcloze} & 205k & Image masking \\
Derain & Graph200k~\cite{li2025visualcloze} & 205k & Raindrop synthesis \\
Low-light enhancement & Graph200k~\cite{li2025visualcloze} & 205k & Physics-based methods \\ \midrule
Editing-Short & OmniEdit~\cite{wei2024omniedit} & 740k &  Mixed expert models \\
Editing-Long & OmniEdit~\cite{wei2024omniedit} & 740k &  Mixed expert models \\ 
Editing-Unique & OmniEdit~\cite{wei2024omniedit} & 410k &  Mixed expert models \\
Style transfer & Stylebooth~\cite{han2025stylebooth} & 11k & De-stylize and restyle \\
    \bottomrule
    \end{tabular}
    }
    \caption{Datamix used to train our \methodname. We form a joint dataset containing 8 million paired data spanning over 15 distinct tasks.}
    \label{tab:supp_data}
\end{table*}

\smallsec{Colorization.} Grayscale images are generated by converting RGB images to LAB space and keeping the L channel.

\smallsec{Inpainting.} We randomly place white or black brush strokes on each clean image. Stroke parameters follow a controlled randomization: we sample between 15–35 strokes and stop early if 30\% of the image area becomes masked. The masked image is used directly as the degraded observation (no separate binary mask input is required).

\smallsec{Derain.} Rain is rendered using a controllable procedural generator. We vary four rain-severity levels -- light, moderate, heavy, and torrential -- while adjusting streak density, streak length, brightness (blending), haze amount, and raindrop occlusion probability. Streaks are implemented via motion-blurred line segments.

\smallsec{Low-light enhancement.} We synthesize under-exposed images by converting inputs to linear space and applying exposure adjustments of $2^{-\text{stop}}$, with $\text{stop}$ randomly sampled in $[3.0, 5.0]$. This yields noise-free but significantly darkened inputs.

\begin{table}[t]
    \centering
    \begin{tabular}{l|cc}
    \toprule
        Config & Stage I & Stage II \\ \midrule
        GPUs & 8 & 8 \\ 
        Batch size & 4 & 2 \\ 
        Grad. accumulation step & 8 & 8 \\ 
        Trainable params & Text projectors & Infinity \\ 
        Training time & 2 epochs & 100k steps \\ 
        Learning rate & 1e-4 & 5e-6 \\ 
        Dropout rate & 0.0 & 0.1 \\ 
        CFG rate & 0.0 & 0.1 \\ 
        Warmup steps & 1000 & 1000 \\ 
        DeepSpeed & Zero1 & Zero1 \\ 
        Optimizer & AdamW & AdamW \\ 
        \bottomrule
    \end{tabular}
    \caption{Training configurations for our \methodname.}
    \label{tab:supp_training}
\end{table}

\smallsec{Denoising (Deblur).} We apply exactly one blur or noise-related degradation per sample, chosen from a small set of common techniques: 
\begin{itemize}
    \item Gaussian blur
    \item Anisotropic Gaussian blur
    \item Motion blur (random camera-shake kernel)
    \item Defocus blur (disk kernel)
    \item Box blur
    \item Median or bilateral smoothing
    \item Downsample–upsample blur
    \item Gaussian noise
    \item Poisson noise
\end{itemize}

\smallsec{Editing.} We use the OmniEdit~\cite{wei2024omniedit} dataset as our training source, which contains 1.2 million source images. Their annotations can be grouped into two categories: data with more than two instructions (short and long versions) and data with a unique instruction. It spans across 6 diverse instruction-guided image editing tasks: Object addition, object removal, and object swap contain both long and short instructions; while attribute editing, environment change, and style transfer contain only one instruction. 

We will release the full synthetic degradation pipeline for reproducibility.

\section{Failure Mode}
\label{suppsec:failure}

\begin{figure}[t]
    \centering
    \includegraphics[width=\linewidth]{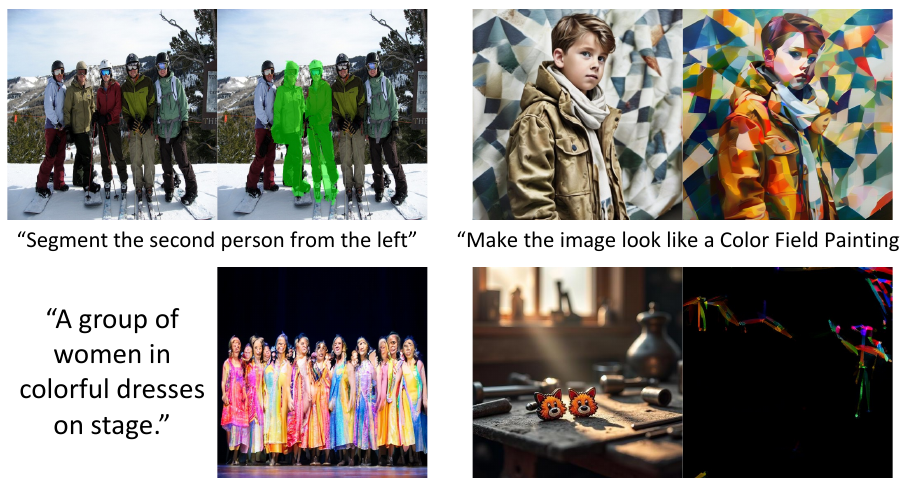}
    \vspace{-10pt}
    \caption{Representative failure mode for our method. Our model has difficulty preserving fine anatomical details; and sometimes meets hallucinations.}
    \label{fig:supp-fail}
    \vspace{-10pt}
    
\end{figure}

While \methodname demonstrates consistently strong performance on image restoration tasks, its limitations become more evident on semantics-heavy or human-centric generation/editing tasks. Figure~\ref{fig:supp-fail} shows typical failure cases. Common issues include: (1) difficulty preserving fine anatomical details, especially under editing-related conditions; (2) similar to prior multi-task UVG models, \methodname may generate outputs that are semantically plausible but not present in the input (example shown bottom-right). Addressing these limitations is an important direction for future work.

\section{Additional Implementation Details}
\label{suppsec:implementation}

\smallsec{Training hyperparameters} for both stages are included in Table~\ref{tab:supp_training}.  We train our model in float32 while offering a mixed precision style inference. During inference, we keep all other parameters in bfloat16, except for a small set of Infinity backbone parameters~\cite{han2025infinity} that are highly sensitive to numerical precision:

\begin{tcolorbox}[colback=white, colframe=black!70, arc=2mm, boxrule=0.5pt]
\textbf{Parameters kept in float32 to enhance pricision}

\texttt{text\_norm} \\
\texttt{text\_proj\_for\_sos} \\
\texttt{text\_proj\_for\_ca} \\
\texttt{shared\_ada\_lin} \\
\texttt{head\_nm} \\
\texttt{head} \\
\texttt{word\_embed} \\
\texttt{norm0\_ve}
\end{tcolorbox}

\smallsec{Reference tokens.} For the tokens from the reference image, we use the raw features from the image encoder instead of using the standard quantized tokens. This design makes sure that our reference tokens contain enough details from the source image, which is critical for UVG tasks.  

\smallsec{System Prompt for Qwen2.5-VL.} We use the same system prompt as Qwen-Image~\cite{wu2025qwen}. Concretely, for the text-to-image (T2I) task, we use:

\begin{tcolorbox}[colback=white, colframe=black!70, arc=2mm, boxrule=0.5pt]
\textbf{System Prompt for T2I task}

\texttt{<|im\_start|>system} \\
Describe the image by detailing the color, quantity, text, shape, size, texture, spatial relationships of the objects and background: \texttt{<|im\_end|>} \\[3pt]
\texttt{<|im\_start|>user} \\
\texttt{<user\_text>} \texttt{<|im\_end|>} \\[3pt]
\texttt{<|im\_start|>assistant}
\end{tcolorbox}

While for other UVG tasks, we use: 

\begin{tcolorbox}[colback=white, colframe=black!70, arc=2mm, boxrule=0.5pt]
\textbf{System Prompt for Other UVG tasks}

\texttt{<|im\_start|>system} \\
Describe the key features of the input image (color, shape, size, texture, objects, background), then
explain how the user's text instruction should alter or modify the image. Generate a new image that
meets the user's requirements while maintaining consistency with the original input where appropriate. \texttt{<|im\_end|>} \\[3pt]
\texttt{<|im\_start|>user} \\
\texttt{<|vision\_start|><|user\_image|>\\
<|vision\_end|>} \\ 
\texttt{<user\_text>} \texttt{<|im\_end|>} \\[3pt]
\texttt{<|im\_start|>assistant}
\end{tcolorbox}

\smallsec{Details about compared models.} 
We report baseline performance directly from the original papers. For our ablation experiments, we additionally train variants built on Qwen2.5-VL~\cite{bai2025qwen2} and Stable Diffusion 3 (SD3)~\cite{esser2024scaling}, adopting the standard SD3 implementation provided in \emph{diffusers}. During inference, we sample using the default setting with 28 steps at a resolution of $512 \times 512$.

\section{Additional Evaluations}
\label{suppsec:rvos}

\subsection{Evaluation on RVS}
Table~\ref{tab:supp-rvos} provides additional quantitative results on the referring segmentation task. We binarize predicted masks and follow the standard RefCOCO evaluation protocol~\cite{yu2016modeling}. Since existing UVG models do not support this task, we compare against specialized methods and include representative LLM-based models for reference.

Our \methodname achieves competitive performance across all three RefCOCO benchmarks. Importantly, UNINEXT~\cite{yan2023universal} is trained using a large mixture of more than 20 datasets, and LISA~\cite{lai2024lisa} relies on pretrained SAM for mask decoding. Despite not using any external segmentation models or additional datasets, \methodname reaches accuracy comparable to LISA and approaches the performance of fully specialized systems. These results highlight the effectiveness of our AR-based unified design and reinforce the potential of autoregressive modeling for UVG tasks that require structured spatial understanding.

\begin{table}[t]
    \centering
    \resizebox{\linewidth}{!}{
    \begin{tabular}{l|ccc|ccc|cc}
    \toprule
    \multirow{2}{*}{Method} & \multicolumn{3}{c|}{RefCOCO} & \multicolumn{3}{c|}{RefCOCO+} & \multicolumn{2}{c}{RefCOCOg} \\ \cline{2-9}
    & val & testA & testB & val & testA & testB & val-u & test-u \\ \midrule
    MCN~\cite{luo2020multi}    &  62.4 & 64.2 & 59.7 & 50.6 & 55.0 & 44.7 & 49.2 & 49.4\\
    VLT~\cite{ding2021vision}     &  67.5 & 70.5 & 65.2 & 56.3 & 61.0 & 50.1 & 55.0 & 57.7\\
    LAVT~\cite{yang2022lavt} & 72.7 & 75.8 & 68.8 & 62.1 & 68.4 & 55.1 & 61.2 & 62.1\\
    LISA-7B~\cite{lai2024lisa} & \underline{74.1} & \underline{76.5} & \underline{71.1} & 62.4 & 67.4 & \underline{56.5} & \underline{66.4} & \underline{68.5}\\ 
    UNINEXT-L~\cite{yan2023universal} & \bf 80.3 & \bf 82.6 & \bf 77.8 & \bf 70.0 & \bf 74.9 & \bf 62.6 & \bf 73.4 & \bf 73.7\\ 
    \midrule 
    \methodname (Ours) & 72.5 & 75.2 & 70.2 & \underline{63.6} & \underline{68.1} & 55.5 & 65.9 & 68.1 \\ 
    \bottomrule
    \end{tabular}
    }
    \caption{Quantitative evaluation on RefCOCO~\cite{yu2016modeling}. Despite not relying on additional segmentation datasets or SAM-based decoders, our unified \methodname achieves competitive results compared to specialized models.}
    \label{tab:supp-rvos}
\end{table}

\subsection{Ablation Study on Training Qwen Encoder}

By default, we set the multimodal encoder as frozen. However, in Table~6 in the main paper, we additionally finetune the last 10 layers of the Qwen model to introduce the additional multimodal understanding task. We additionally ablate the effect of the trainable parameters of the Qwen model in this section. Concretely, we train an additional baseline in which we also finetune the last 10 layers of the Qwen encoder \textit{without} the auxiliary multimodal objective, while keeping all the training recipe identical. Under this setting, the GenEval score drops from \textbf{0.64} to \textbf{0.62}, indicating that the additionally trainable parameters in Qwen do not bring improvement for UVG. This result explains that the gain in Table 6 is due to the multimodal understanding supervision. We interpret this performance drop as a similar issue we discussed in Section 4.3, where fine-tuning large pretrained models on smaller or distribution-shifted datasets can lead to performance drops due to catastrophic forgetting or overfitting.  

\subsection{Ablation Study on Task Mixing ratio}

In the main paper, we set the ratio of T2I batches as 0.25. Here we ablate this design. In Table~\ref{tab:supp-ratio}, we observe that there is a tradeoff controlled by the T2I sampling ratio: too little T2I data harms GenEval, while too much T2I data can degrade restoration. Overall, performance is not highly sensitive to the mixing ratio; varying the ratio within a reasonable range leads to similar results, and our default choice of 0.25 provides a robust balance across tasks.

\begin{table}[t]
    \centering
    \resizebox{\linewidth}{!}{
    \begin{tabular}{l|cccc}
    \toprule
        T@I Ratio & GenEval-Overall ($\uparrow$) & LOL-PSNR ($\uparrow$)  & GoPro-PSNR ($\uparrow$) & Rain100L-PSNR ($\uparrow$) \\ \midrule 
        0.1 & 0.63  &  \bf 21.58  & \bf 23.10 & 33.46  \\
        0.25 & \bf 0.68 & 21.03 &  22.99 & \bf 33.71 \\
        0.5 & \bf 0.68 & 19.98 & 22.53 & 32.80\\
    \bottomrule
    \end{tabular}
    }
    \caption {Trade-off of the data mix ratio: too little T2I data harms GenEval, while too much T2I data can degrade restoration. Varying the ratio within a reasonable range leads to similar results.}
    \label{tab:supp-ratio}
    \vspace{-10pt}
\end{table}

\subsection{Ablation Study on Design of Ref Tokens}

In this section, we ablate the usage of the reference tokens. By default, we only select the highest resolution tokens as the reference tokens. We follow the protocol of EditVAR~\cite{mao2025visual} and design a version using full-scale reference tokens, and the results are reported in Table~\ref{tab:supp-token}. Consistent with prior findings, using full reference tokens does not yield additional performance improvements while increasing inference latency. 

\begin{table}[t]
    \centering
    \resizebox{\linewidth}{!}{
    \begin{tabular}{l|ccc|c}
    \toprule
    Ref Token Design &  LOL-PSNR ($\uparrow$) & GoPro-PSNR ($\uparrow$) & Rain100L-PSNR ($\uparrow$) & Inf. time (s/img) \\
        \midrule
    Finest & \bf 21.03 & \bf 22.99 & \bf 33.71 & \bf 1.05 \\ 
    Full & 20.62 & 22.15 & 31.08 & 1.57 \\
    \bottomrule
    \end{tabular}
    }
    \caption{Comparison with full-scale reference tokens. Using full reference tokens does not yield additional performance improvements while increasing inference latency.}
    \label{tab:supp-token}
    \vspace{-10pt}
    
\end{table}

\begin{figure*}[t]
\centering
\includegraphics[width=\linewidth]{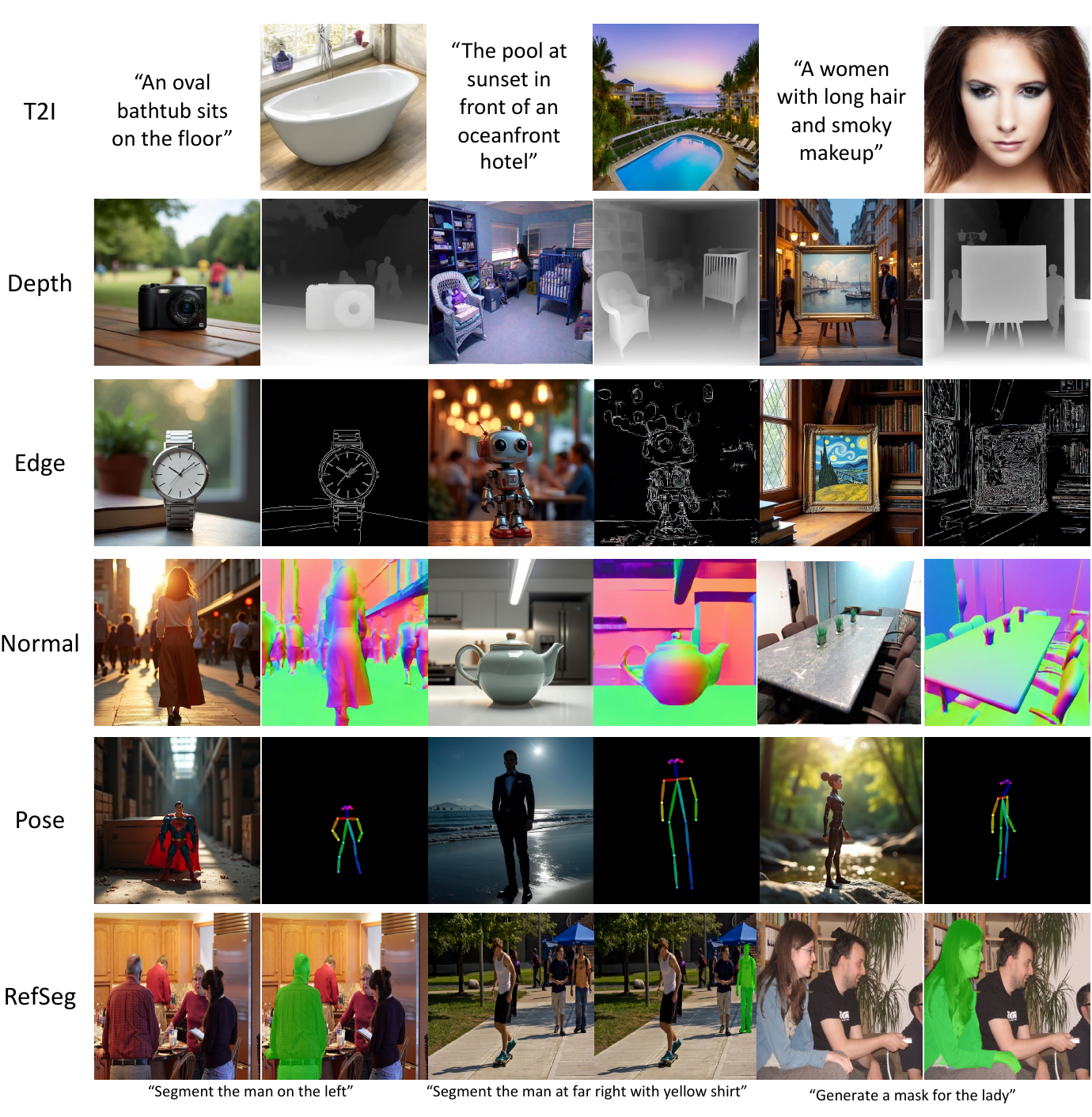}
\vspace{-15pt}
\caption{Additional qualitative results for text-to-image generation and perception tasks. \methodname demonstrates strong generalization.}
\label{fig:supp-cv}
\end{figure*}

\section{Additional Visualizations}
\label{suppsec:visual}

\begin{figure*}[t]
\centering
\includegraphics[width=\linewidth]{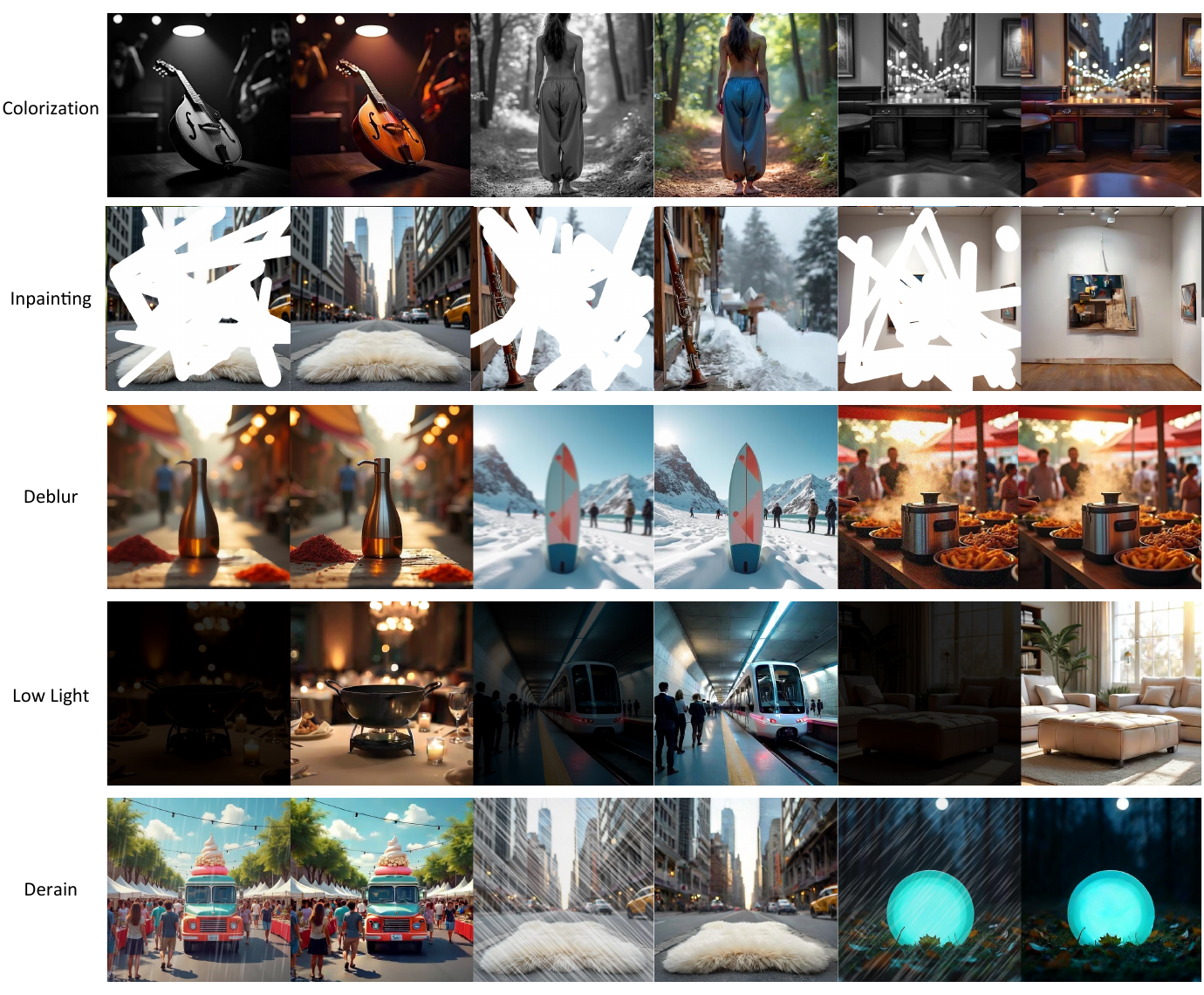}
\caption{Additional qualitative results for image restoration tasks. Our \methodname achieves high-quality outputs across a wide range of restoration settings.}
\label{fig:supp-restore}
\end{figure*}

\begin{figure*}[t]
\centering
\includegraphics[width=\linewidth]{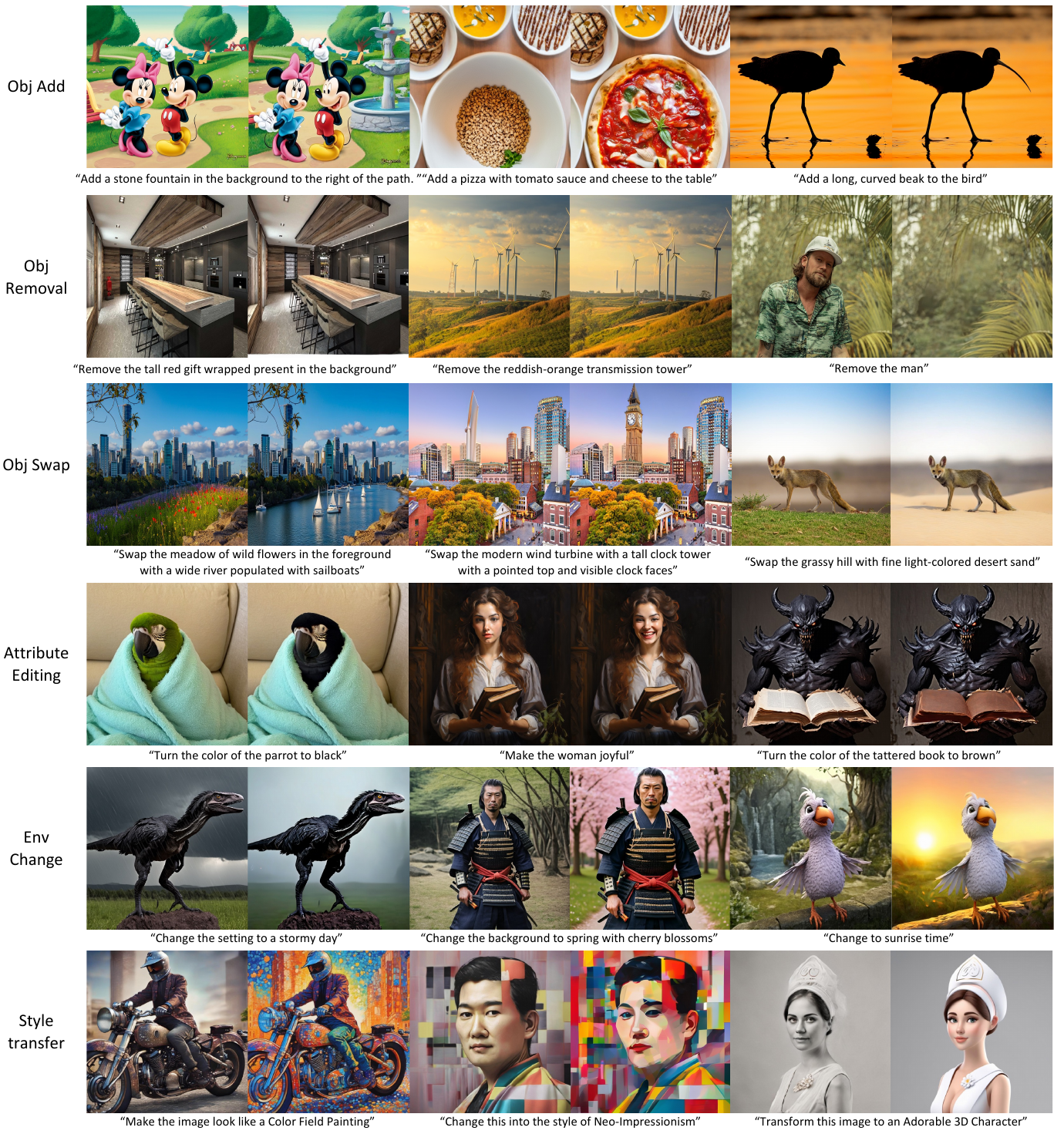}
\caption{Additional qualitative results for image editing tasks. Our \methodname demonstrates superior performance on a diverse set of editing tasks.}
\label{fig:supp-editing}
\end{figure*}

We include additional qualitative results for all tasks considered in the main paper.
Figure~\ref{fig:supp-cv} shows examples of text-to-image generation and perception tasks, Figure~\ref{fig:supp-restore} presents extended results for image restoration, and Figure~\ref{fig:supp-editing} presents additional visualizations for editing tasks. These examples further illustrate the broad generalization ability of \methodname across both perception-centric and editing-centric UVG tasks.

\end{document}